\documentclass{article}

\usepackage{PRIMEarxiv}

\usepackage[utf8]{inputenc} 
\usepackage[T1]{fontenc}    
\usepackage{hyperref}       
\usepackage{url}            
\usepackage{booktabs}       
\usepackage{amsfonts}       
\usepackage{nicefrac}       
\usepackage{microtype}      
\usepackage{fancyhdr}       
\usepackage{graphicx}       
\graphicspath{{media/}}     

\usepackage{amssymb}
\usepackage{amsmath}
\usepackage{booktabs}
\usepackage{tabularx}
\newcolumntype{Y}{>{\centering\arraybackslash}X}
\usepackage{xcolor}
\usepackage{graphicx}

\definecolor{vulnerable}{HTML}{dc3545}
\definecolor{almost_vulnerable}{HTML}{fd7e14}
\definecolor{almost_immune}{HTML}{ffc107}
\definecolor{immune}{HTML}{28a745}

\title{Federated Knowledge Recycling:\\ Privacy-Preserving Synthetic Data Sharing
}

\author{
  Eugenio Lomurno, Matteo Matteucci \\
  Politecnico di Milano \\
  Department of Electronics, Information and Bioengineering\\
  Via Ponzio 34/5, 20133 Milan, Italy\\
  \texttt{\{eugenio.lomurno, matteo.matteucci\}@polimi.it} \\
}

\begin{document}
\maketitle

\begin{abstract}
\noindent Federated learning has emerged as a paradigm for collaborative learning, enabling the development of robust models without the need to centralise sensitive data. However, conventional federated learning techniques have privacy and security vulnerabilities due to the exposure of models, parameters or updates, which can be exploited as an attack surface. This paper presents Federated Knowledge Recycling (FedKR), a cross-silo federated learning approach that uses locally generated synthetic data to facilitate collaboration between institutions. FedKR combines advanced data generation techniques with a dynamic aggregation process to provide greater security against privacy attacks than existing methods, significantly reducing the attack surface. Experimental results on generic and medical datasets show that FedKR achieves competitive performance, with an average improvement in accuracy of 4.24\% compared to training models from local data, demonstrating particular effectiveness in data scarcity scenarios.
\end{abstract}


\section{Introduction}\label{sec:Introduction}
\noindent Deep learning has made significant progress in recent years, with applications in many areas of society. 
Generative deep learning, which can produce synthetic data of very high quality regardless of its type and from different forms of conditioning, is currently one of the most promising subfields. At the same time, federated learning is emerging as a paradigm to enable collaboration between multiple realities, balancing the need to feed information into increasingly powerful models with the need to protect privacy and provide robust guarantees to users.

\noindent Federated learning is particularly relevant in the health sector, where it is often complex to create datasets of sufficient size in a single location. In this context, it is referred to as cross-silo federated learning, as the members of the federation are centres of aggregation of end-user data, in this case individual patients. The primary goal is to maintain the highest level of privacy for such highly sensitive data by using techniques to share information without actually exposing the data to the public.
However, conventional federated learning techniques, while solving the problem of direct data exchange, expose other types of information such as weights, gradients or model logits, effectively creating new vulnerable attack surfaces. 

\noindent To deal with such countermeasures, numerous protection techniques have been proposed, each designed ad hoc to address one or more specific vulnerabilities. While in many cases these techniques are successful in preserving privacy, they often excessively compromise the performance of the federation, either by creating private but overly inefficient models, or by requiring prohibitively long computation times.

\noindent In response to these challenges, this paper presents a cross-silo federated learning technique called Federated Knowledge Recycling (FedKR). This methodology aims to mitigate most of the privacy risks of current federated systems through the use of generative models and the exchange of only synthetic data, with the goal of preserving the privacy of the data and models involved in the federation process, while maintaining beneficial performance for the members of the federation.
The FedKR system requires each participant, in order to join the federation, to provide a set of synthetic data related to a particular type of problem to be solved or category of data. This mechanism acts as a membership to the federation and allows access to the synthetic data produced by all other users and stored in a Central Server, creating a pool of shared knowledge without compromising the confidentiality of the actual original data.

\noindent Finally, through a tuning step to select the most performing synthetic datasets, called Dynamic Dataset Aggregation, each user can locally build their own aggregated dataset and train the optimal model with respect to their own data. This approach eliminates the need to publicly expose private data or models, and offers superior properties and privacy guarantees compared to conventional federated learning methods.
Thus, FedKR is proposed as a solution that combines the advantages of inter-institutional collaboration with robust privacy guarantees. By means of simulations on medical datasets of different types and sizes, the effectiveness of this approach in areas such as healthcare is demonstrated in terms of privacy and the development of shared models that perform better than those otherwise obtainable through non-collaboration.

\noindent The main contributions of this research are:
\begin{itemize}
    \item The introduction of Federated Knowledge Recycling (FedKR), an innovative federated learning paradigm that uses only synthetic data to exchange information between participants, ensuring a high level of privacy for the original data and reducing exposure to privacy threats compared to existing techniques.
    \item The proposal of the Dynamic Dataset Aggregation technique within FedKR, which allows dynamic optimisation of shared synthetic datasets, enabling each federation member to create a customised training dataset and further reducing the privacy attack surface.
    \item The experimental validation of FedKR in the field of medical imaging, demonstrating its effectiveness in balancing the protection of sensitive data with the need to develop high-performance models, especially in contexts of limited data availability.
\end{itemize}

\section{Related Works}\label{sec:Related}
\noindent Federated learning is a collaborative learning technique that aims to centralise information in order to develop more effective models than those created individually, while preserving the privacy of the data used in training. 
Over time, several strategies have emerged to refine this methodology. 
Federated Averaging (FedAvg) is the most widely used model in federated learning. This approach aggregates the parameters or gradients of local model updates through averaging to create a global model. The process involves training clients locally, sending the updates to a central server, and redistributing the aggregated model~\cite{mcmahan2017communication}.
FedProx extends FedAvg and introduces a proximity term to handle client heterogeneity, limiting the divergence between local and global models~\cite{li2020federated}. 
SCAFFOLD addresses client drift by using control variables to correct and guide the flow of updates, improving convergence in the presence of heterogeneous data~\cite{karimireddy2020scaffold}. 
FedNova, designed for non-IID data scenarios, implements a normalisation mechanism that balances the contribution of each client and adjusts local updates according to different amounts of data and iterations~\cite{wang2020tackling}.

\noindent Recently, the use of synthetic data in federated learning has gained interest as an alternative solution to traditional methods to improve privacy and maintain efficient learning.
FedGAN presents an approach where each participant trains a GAN locally on its own data. The local datasets are enriched by mixing real and generated data, which are then used in a process similar to that proposed in FedAvg. This method aims at preserving privacy by introducing uncertainty in the real data~\cite{rasouli2020fedgan}.
Other variants of the same approach aim to use generators with differential privacy guarantees, and instead of building a mixed dataset of real and synthetic data, they send only the latter to the central server for training the global model~\cite{augenstein2019generative}.
FedMatch, on the other hand, is a technique based on a semi-supervised approach using a pseudo-label generator. Clients locally generate pseudo-labels for unlabelled data, which are then used in a federated learning process. This method uses synthetic pseudo-labelled data to improve performance in scenarios with limited labelled data~\cite{jeong2021federated}.
Instead, FedSyn is a method that uses a shared generator to produce synthetic data representing the global distribution. Each client uses this synthetic data to regularise its own local~\cite{li2022federated} model.
SGDE, on the other hand, is a federated system in which each user uses its own private data to locally train Variational Autoencoder-type generators with differential privacy guarantees. The generators are then sent to a central server where all other shared generators can be accessed to locally generate their own synthetic dataset and train their own model~\cite{lomurno2022sgde}.

\subsection{Threats and Prevention for Federated Learning}
\noindent Although federated learning originated as a paradigm aimed at preserving privacy while sharing information to build better performing models, the sharing of such information or models creates a very large attack surface. Some of the most common and studied attacks include:
\begin{itemize}
    \item \textbf{Model Inversion}. This attack aims to reconstruct training data from common model parameters. By analysing these parameters, an attacker can potentially infer significant details about the original data used for training~\cite{fredrikson2015model}.
    \item \textbf{Membership Inference}. The aim is to determine whether a specific sample of data was used to train the model. By analysing how the model responds to certain inputs, an attacker can determine with some probability the presence of certain data in the training set~\cite{shokri2017membership}.
    \item \textbf{Data Poisoning}. These attacks aim to degrade the performance of the global model by injecting malicious or modified data into the local training process. This can be aimed at introducing biases or compromising overall model accuracy~\cite{tolpegin2020data}.
    \item \textbf{Byzantine}. These attacks aim to sabotage the aggregation process by sending arbitrary or malicious updates from some clients, thus compromising the convergence of the training and the quality of the final model~\cite{yin2018byzantine}.
    \item \textbf{Sybil}. The Sybil attack manipulates the aggregation process by creating multiple false identities, artificially increasing the attacker's influence over the final model. This can lead to disproportionate control over the training result~\cite{wang2020attack}.
    \item \textbf{Evasion}. Evasion attacks aim to fool the model during inference through adversarial inputs specifically designed to cause misclassifications. These attacks are particularly effective in bypassing security systems such as malware detection, manipulating automated decisions in autonomous vehicles, or compromising medical diagnoses~\cite{biggio2013evasion}.
    \item \textbf{Gradients/Weights Leakage}. Similar to model inversion, this attack aims to reconstruct training data by analysing the weights or gradients shared during the federated model update process~\cite{zhu2019deep}.
    \item \textbf{Free-Riding}. This attack occurs when participants benefit from the global model without contributing significantly to its improvement by sending false or low quality updates. This can lead to a degradation of overall system performance~\cite{fraboni2021free}.
    \item \textbf{Property Inference}. These attacks aim to infer statistical properties of the training dataset by analysing the behaviour of the model on different inputs. The goal is to derive aggregate information about the data without directly accessing it~\cite{ganju2018property}.
    \item \textbf{Temporal}. Temporal attacks exploit differences in response or update times to derive information about the hardware or data of the participants. This information can be used to identify vulnerabilities or specific characteristics of the devices involved~\cite{nasr2019comprehensive}.
\end{itemize}

\noindent These threats are constantly evolving in number and effectiveness, but this growth has also led to the development of sophisticated countermeasures that are effective but come at the cost of some performance. 
Differential privacy emerges as a key technique against membership inference and model inversion attacks, adding calibrated noise to data or models to protect individual information. However, the main trade-off is a potential reduction in model accuracy, with the magnitude of the trade-off depending on the desired level of privacy~\cite{abadi2016deep,jordon2018pate}. Recently, other strategies based on regularisation or adversarial training have been tested to prevent the same attacks and achieve a better trade-off between accuracy and resistance to membership inference and model inversion~\cite{lomurno2022utility,lomurno2023discriminative}.
Homomorphic encryption, which is effective against data leakage attacks during computation, allows processing on encrypted data, providing robust protection. The key trade-off is the introduction of significant computational overhead, which can significantly slow down the training process~\cite{acar2018survey}.
Secure aggregation techniques protect privacy during model aggregation by counteracting Gradients/Weights Leakage attacks. These techniques balance security and efficiency, but may introduce latency in the global model update process~\cite{bonawitz2017practical}.
Approaches such as model pruning and quantisation reduce the attack surface for model inversion attacks by limiting the information shared. However, these methods can lead to potential losses in model performance, especially for complex tasks~\cite{wang2018atomo}.

\section{Method}\label{sec:Method}
\noindent This section introduces the Federated Knowledge Recycling (FedKR) technique, an alternative approach to current federated learning methods that aims to provide high privacy guarantees through the use of generated data. This technique is applied and tested in a federated scenario where medical images are shared for classification tasks.
The core principle of FedKR is based on the process of sharing and aggregating exclusively synthetic data. To participate in the federation, each member is required to generate a synthetic dataset using a generator trained locally on its own private data. This dataset is then sent to a central server, which acts as a shared repository and distribution hub for the entire federation.

\begin{figure}[t]
    \centering
    \includegraphics[width=1.\linewidth]{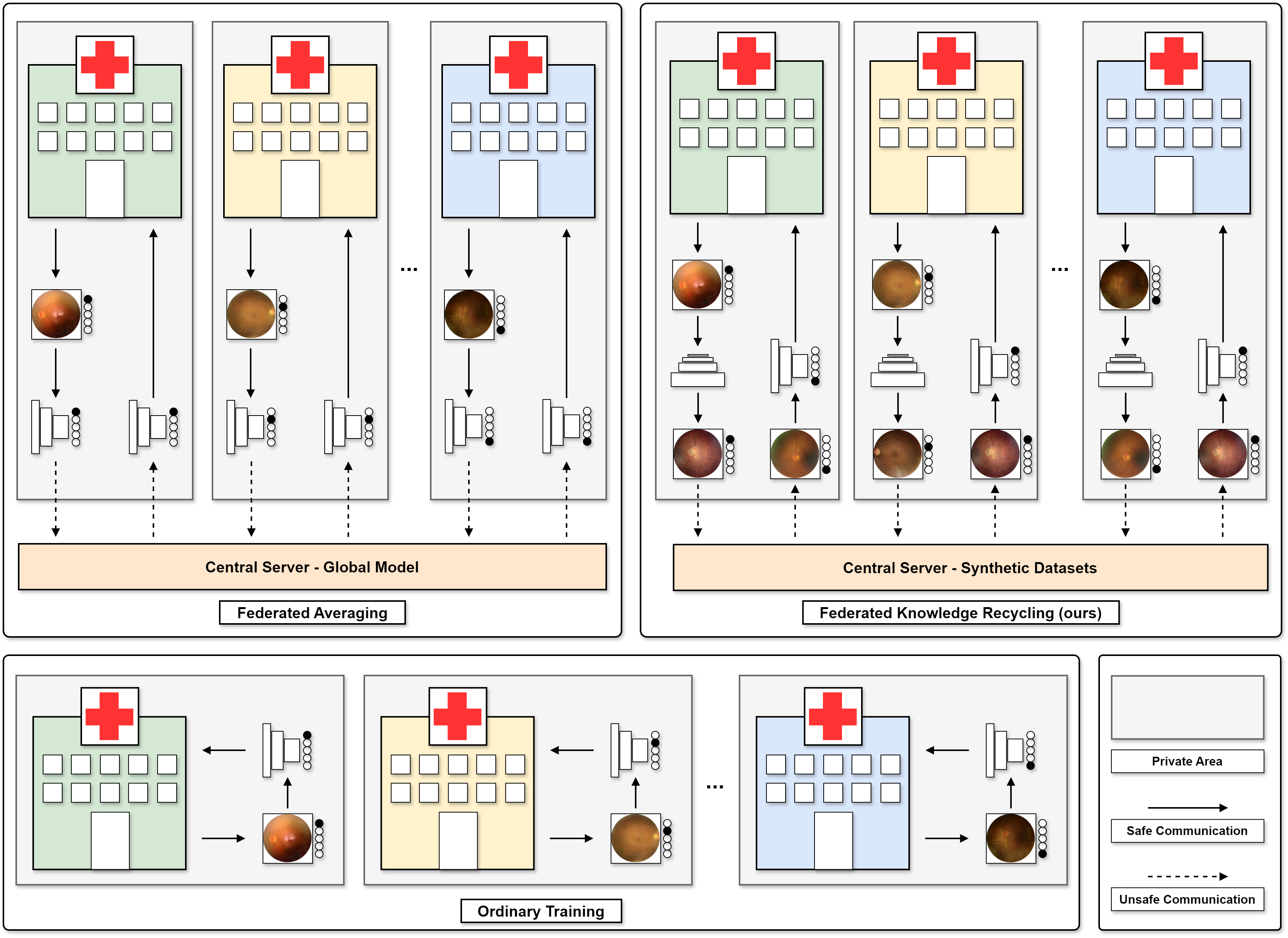}
    \caption{\footnotesize{The difference between an Ordinary Training, Federated Averaging and the Federated Knowledge Recycling technique.}}
    \label{fig:Pipeline}
\end{figure}

\noindent In return for sharing their own synthetic dataset, each participant gains access to synthetic datasets provided by other members. These can be downloaded locally, allowing each user to construct their own customised aggregated synthetic dataset. On this aggregated basis, the user can then train their model in a fully autonomous and localised manner.
This thesis also presents a technique for managing and optimising synthetic datasets called Dynamic Datasets Aggregation. This technique makes it possible to dynamically modulate the contribution of each dataset within the final aggregated dataset. This approach offers the possibility to optimise the performance of the model by giving more weight to the most relevant datasets and, if necessary, excluding the less relevant ones from the final aggregated dataset. A graphical representation is shown in Figure\ref{fig:Pipeline}.

\subsection{Knowledge Recycling}
\noindent To enable an exchange of synthetic data that provides useful information for machine learning tasks to other members of the federation, specific generation and post-processing methods are required. A growing number of techniques and strategies have been proposed in the literature to generate synthetic datasets to replace real ones~\cite{dat2019classifier,lampis2023bridging,lomurno2024stable}.
In this work, each member of the federation uses the Knowledge Recycling, a pipeline for the creation of synthetic datasets and their subsequent use in classifier training, designed to maximise the usefulness of such data and make the models trained with it highly resilient to Membership Inference attacks~\cite{lomurno2024synthetic}.
As in the reference work, private data is used exclusively to train a Generator and a Teacher Classifier. The whole pipeline aims at obtaining the best Student Classifier, a model identical in structure and training to the Teacher Classifier, but trained only on synthetic data generated by the Generator. A modified version of the BigGAN-Deep model is used for the Generator, while a ResNet14 architecture is used for the Teacher and Student Classifiers. For implementation and training details, the reader is referred to the reference article.

\noindent Throughout the whole process, from the selection of the best checkpoint to the final generation of the dataset, the Generative Knowledge Distillation technique -- presented in the same paper -- is applied. In this technique, the Teacher Classifier evaluates the synthetic images and generates soft labels, i.e. probability distributions over the classes, instead of binary labels. These soft labels capture nuances and uncertainties, thus adding richer knowledge to the synthetic dataset.
The process starts with identifying the best checkpoint of the Generator based on the Classification Accuracy Score (CAS) metric, which indicates the accuracy performance of a classifier trained on generated data and evaluated on real data~\cite{ravuri2019classification}. Next, a 50-step tuning step is performed to determine the optimal generation standard deviation, again to maximise the CAS of the Student Classifier. This optimisation is performed using a Tree-structured Parzen Estimator combined with a Hyperband pruning mechanism~\cite{bergstra2011algorithms,li2018hyperband}. The range of allowable values is from 0.5 to 2.5. Once all these parameters have been established, the Generator is used to create the final synthetic dataset, producing images that mimic the distribution of the real data. A synthetic dataset five times the size of the real one is generated and sent to the Central Server.

\subsection{Dynamic Datasets Aggregation}
\noindent The final step of the FedKR technique, called Dynamic Datasets Aggregation (DDA), involves the access to and use of the synthetic datasets of the whole federation for the local training. The process starts with the local download of the synthetic datasets of interest. Next, the aggregation of these datasets is optimised through a 50-step process using a Tree-structured Parzen Estimator combined with a Hyperband pruning mechanism, similar to the previous step.

\noindent During this optimisation step, two key parameters are calibrated. The first is the percentage contribution of each synthetic dataset with respect to its original size. This allows the contribution of each dataset to be finely modulated, from zero to 100\% of its data, with a granularity of 20\%. The second parameter is the regeneration rate of the aggregated dataset during the training of the classifier.
The regeneration rate determines how the aggregated dataset is used during training. This can vary from one regeneration at each epoch to a single generation at the start of training. If the rate is set to regenerate only at the start of training, the entire aggregate dataset will be used in one go. Otherwise, depending on the set number of regenerations, the aggregated dataset is divided into equivalent parts, exploiting the abundance of synthetic data available. Both parameters are optimised to maximise the local CAS.
The classifier trained in this way is always the same ResNet14 model used in the previous steps~\cite{lomurno2024synthetic}. Training is performed for a total of 100 epochs, regardless of the regeneration rate chosen.

\begin{figure}[p]
    \centering
    \includegraphics[width=0.7\linewidth]{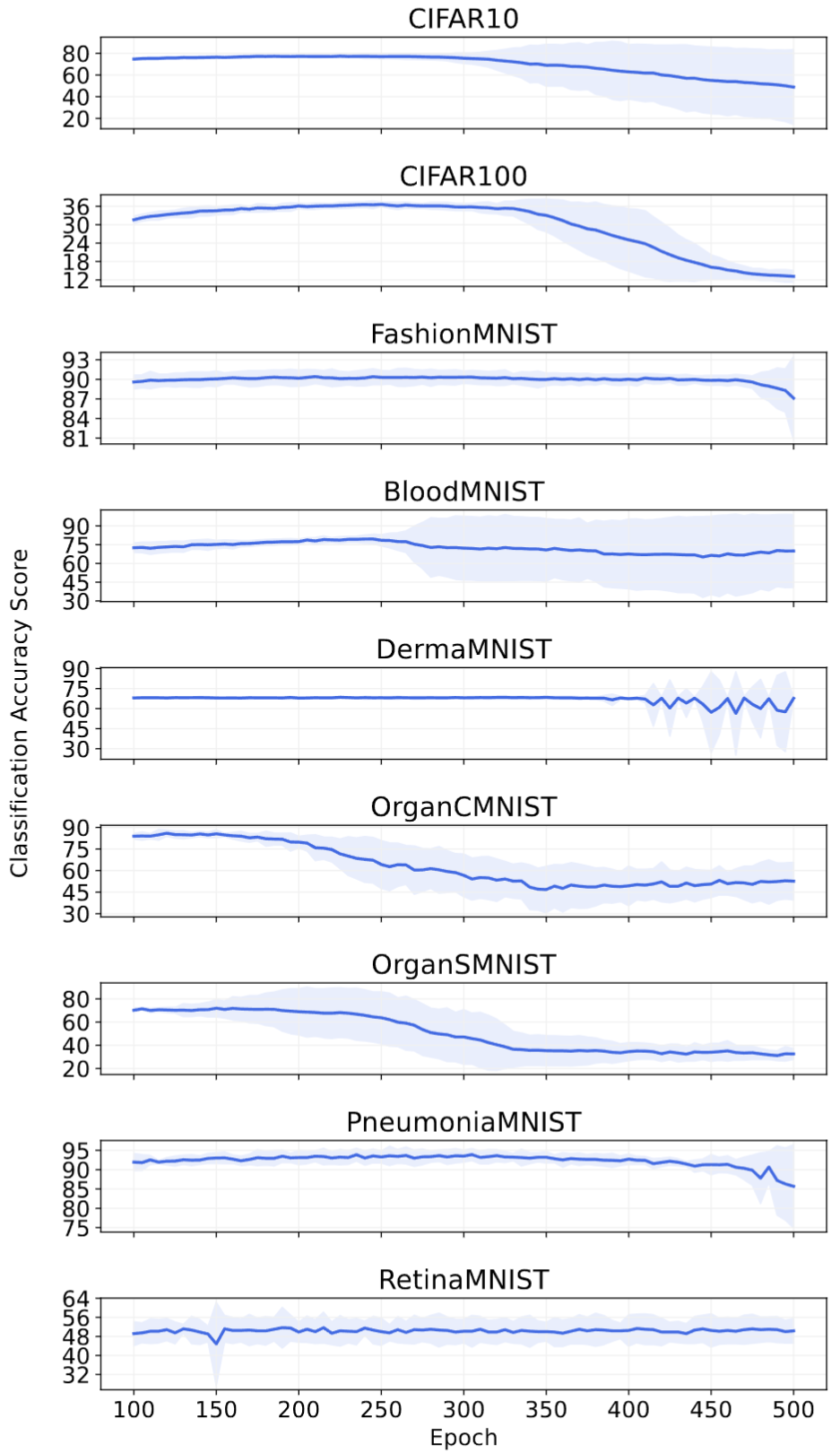}
    \caption{\footnotesize{The performance achieved by the members of the federation in the identification of the optimal checkpoint of the Generator to be shared.}}
    \label{fig:checkpoint}
\end{figure}

\section{Results and Discussion}\label{sec:Results}
\noindent The experiments were conducted on nine image datasets, all rescaled to 32x32 pixels. The first three generic datasets -- CIFAR10, CIFAR100 and FashionMNIST -- were used both for final comparisons and to calibrate and test the proposed Federated Knowledge Recycling (FedKR) technique~\cite{krizhevsky2009learning,xiao2017fashion}. The next six are medical datasets -- BloodMNIST, DermaMNIST, OrganCMNIST, OrganSMNIST, PneumoniaMNIST and RetinaMNIST -- and contain real images from the MedMNIST v2 benchmark~\cite{medmnistv2}. These medical datasets represent the main focus of the proposed technique, once calibrated on the three previous datasets.
The experimental setup simulated a federated cross-silo scenario with 20 members. The training, validation and test sets were divided into 20 equal parts. The samples were randomly distributed with no class stratification. The experiments were run on 4 NVIDIA Quadro RTX 6000 GPUs.

\noindent Within the simulation, each federation member trains its own Generator, identifies the best checkpoint and standard deviation with respect to its validation data, and generates the synthetic dataset to be shared, which is sent to the Central Server. 
Details on the performance of the Generators' checkpoints can be found in Figure~\ref{fig:checkpoint}, where it is possible to see both the high level of Classification Accuracy Score achieved by each federation member on its actual validation data, and the heterogeneity of this performance due to the unstratified sharing of the data.
Next, each federation member downloads each of the 20 synthetic datasets to the Central Server and performs the Dynamic Dataset Aggregation step with all of them.

\noindent The proposed FedKR technique is compared to FedAvg, which is implemented without the use of privacy attack prevention techniques. The comparison also includes Ordinary Training, which is used as a lower bound and represents the training that each federation member could perform as an alternative to joining the federation. Centralised Training simulates the scenario where each member would share its private data to train a global model, providing a theoretical upper bound on achievable performance that is clearly not feasible in practice unless the private data is publicly exposed.
For robustness of comparison, each ResNet14 classifier used was trained for 100 epochs with the same training technique regardless of the scenario~\cite{lomurno2024synthetic}.

\begin{table*}[t]
    \centering
    \setlength{\tabcolsep}{3pt}
    \caption{\footnotesize{Comparison of test Accuracy achieved by proposed approaches -- for FedKR approach, Accuracy is intended as Classification Accuracy Score.}}
    \vspace{4pt}
    \footnotesize
    \begin{tabularx}{\textwidth}{l*{4}{Y}}
        \toprule
        \footnotesize{Model} & \footnotesize{Ordinary Training} & \footnotesize{Centralised Training} & \footnotesize{FedAvg} & \footnotesize{FedKR (ours)}\\
        \midrule
        \footnotesize{CIFAR10}         &  72.20 $\pm$ 3.06    &   94.95 $\pm$ 0.52     &   86.95 $\pm$ 0.55    &   80.16 $\pm$ 0.81\\
        \footnotesize{CIFAR100}        &  27.71 $\pm$ 3.19    &   75.70 $\pm$ 1.07     &   55.75 $\pm$ 2.16    &   36.60 $\pm$ 1.26\\
        \footnotesize{FashionMNIST}    &  89.24 $\pm$ 2.38    &   95.54 $\pm$ 0.49     &   93.43 $\pm$ 1.30    &   91.68 $\pm$ 0.71\\
        \midrule[0.05pt]
        \footnotesize{BloodMNIST}      &  90.19 $\pm$ 4.26    &   97.93 $\pm$ 0.47     &   95.79 $\pm$ 0.77    &   90.91 $\pm$ 1.03\\
        \footnotesize{DermaMNIST}      &  67.92 $\pm$ 5.26    &   79.40 $\pm$ 1.98     &   73.46 $\pm$ 2.87    &   70.47 $\pm$ 2.42\\
        \footnotesize{OrganCMNIST}     &  80.68 $\pm$ 3.63    &   93.29 $\pm$ 0.59     &   88.95 $\pm$ 0.80    &   83.19 $\pm$ 0.62\\
        \footnotesize{OrganSMNIST}     &  69.45 $\pm$ 3.31    &   82.64 $\pm$ 0.71     &   78.05 $\pm$ 0.88    &   70.19 $\pm$ 0.99\\
        \footnotesize{PneumoniaMNIST}  &  79.25 $\pm$ 9.30    &   83.65 $\pm$ 4.09     &   88.72 $\pm$ 6.04    &   89.59 $\pm$ 2.47\\
        \footnotesize{RetinaMNIST}     &  49.00 $\pm$ 15.45   &   53.75 $\pm$ 4.55     &   55.99 $\pm$ 11.47   &   50.99 $\pm$ 6.64\\
        \midrule
        \footnotesize{Min Imp}         &  -       &   4.39      &   4.19    &   0.72\\
        \footnotesize{Mean Imp}        &  -       &   14.50     &   10.10   &   4.24\\
        \footnotesize{Max Imp}         &  -       &   47.70     &   28.04   &   10.33\\
        \bottomrule
    \end{tabularx}
    \label{tab:Performance}
\end{table*}

\begin{table*}[!]
    \centering
    \setlength{\tabcolsep}{6pt}
    \caption{\footnotesize{The privacy attack resistance properties of the examined approaches compared to the considered attacks. The red dot (\textcolor{vulnerable}{$\bullet$}) indicates total vulnerability, the orange dot (\textcolor{almost_vulnerable}{$\bullet$}) near vulnerability, the yellow dot (\textcolor{almost_immune}{$\bullet$}) near immunity, and the green dot (\textcolor{immune}{$\bullet$}) immunity. The dash (-) indicates that the attack is not applicable. Each rating reflects the ability of an attack to compromise the privacy of real private data using that attack technique.}}
    \vspace{4pt}
    \footnotesize
    \begin{tabularx}{\textwidth}{l*{4}{Y}}
        \toprule
        \footnotesize{Attack} & \footnotesize{Ordinary Training} & \footnotesize{Centralised Training} & \footnotesize{FedAvg} & \footnotesize{FedKR (ours)}\\ 
        \midrule
        \footnotesize{Model Inversion}              & \textcolor{vulnerable}{\scalebox{1.75}{$\bullet$}} & \textcolor{vulnerable}{\scalebox{1.75}{$\bullet$}} & \textcolor{vulnerable}{\scalebox{1.75}{$\bullet$}} & \textcolor{immune}{\scalebox{1.75}{$\bullet$}}\\
        \footnotesize{Membership Inference}         & \textcolor{vulnerable}{\scalebox{1.75}{$\bullet$}} & \textcolor{vulnerable}{\scalebox{1.75}{$\bullet$}} & \textcolor{vulnerable}{\scalebox{1.75}{$\bullet$}} & \textcolor{immune}{\scalebox{1.75}{$\bullet$}}\\
        \footnotesize{Data Poisoning}               & - & \textcolor{almost_vulnerable}{\scalebox{1.75}{$\bullet$}} & \textcolor{vulnerable}{\scalebox{1.75}{$\bullet$}} & \textcolor{almost_immune}{\scalebox{1.75}{$\bullet$}}\\
        \footnotesize{Byzantine}                    & - & - & \textcolor{vulnerable}{\scalebox{1.75}{$\bullet$}} & \textcolor{almost_immune}{\scalebox{1.75}{$\bullet$}}\\
        \footnotesize{Sybil}                        & - & - & \textcolor{vulnerable}{\scalebox{1.75}{$\bullet$}} & \textcolor{almost_immune}{\scalebox{1.75}{$\bullet$}}\\
        \footnotesize{Evasion}                      & \textcolor{almost_vulnerable}{\scalebox{1.75}{$\bullet$}} & \textcolor{almost_vulnerable}{\scalebox{1.75}{$\bullet$}} & \textcolor{almost_vulnerable}{\scalebox{1.75}{$\bullet$}} & \textcolor{almost_vulnerable}{\scalebox{1.75}{$\bullet$}}\\
        \footnotesize{Gradients/Weights Leakage}    & \textcolor{vulnerable}{\scalebox{1.75}{$\bullet$}} & \textcolor{vulnerable}{\scalebox{1.75}{$\bullet$}} & \textcolor{vulnerable}{\scalebox{1.75}{$\bullet$}} & \textcolor{immune}{\scalebox{1.75}{$\bullet$}}\\
        \footnotesize{Free-Riding}                  & - & - & \textcolor{almost_vulnerable}{\scalebox{1.75}{$\bullet$}} & \textcolor{almost_immune}{\scalebox{1.75}{$\bullet$}}\\
        \footnotesize{Property Inference}           & \textcolor{almost_vulnerable}{\scalebox{1.75}{$\bullet$}} & \textcolor{vulnerable}{\scalebox{1.75}{$\bullet$}} & \textcolor{almost_vulnerable}{\scalebox{1.75}{$\bullet$}} & \textcolor{almost_immune}{\scalebox{1.75}{$\bullet$}}\\
        \footnotesize{Temporal}                     & \textcolor{immune}{\scalebox{1.75}{$\bullet$}} & \textcolor{immune}{\scalebox{1.75}{$\bullet$}} & \textcolor{almost_vulnerable}{\scalebox{1.75}{$\bullet$}} & \textcolor{almost_immune}{\scalebox{1.75}{$\bullet$}}\\
        \bottomrule
    \end{tabularx}
    \label{tab:Privacy}
\end{table*}

\noindent The results, presented in Table~\ref{tab:Performance}, show that participation in a federation using FedKR leads to an average improvement in accuracy on test data of 4.24\% compared to Ordinary Training. The improvement ranges from a minimum of 0.72\% for BloodMNIST to a maximum of 10.33\% for PneumoniaMNIST, demonstrating the effectiveness of both the quality of generation and the application of Dynamic Dataset Aggregation.
Centralised Training generally achieves the best performance, but remains an ideal case that is not applicable for privacy reasons. It is interesting to note the case of the PneumoniaMNIST dataset, with only 235 training samples per federation member, where FedKR manages to outperform all other techniques, including Centralised Training. This outlier case demonstrates the potential benefits of synthetic data abundance under conditions of data scarcity, a common situation in the clinical context, e.g. for rare diseases.
FedAvg proves to be a high performance federated system, with an average accuracy improvement of 10.10\% over Ordinary Training, but has significant privacy limitations.

\noindent The Table~\ref{tab:Privacy} compares the resilience properties to various privacy attacks without applying any countermeasures to the techniques compared to mitigate their effects. This simplifying choice stems from the fact that these mitigations are largely applicable in all scenarios with appropriate adaptations.
Furthermore, although the main comparison models are FedAvg and FedKR, for the sake of completeness, the resilience levels of Ordinary Training -- in the case where one decides to share locally trained models on one's own private data -- and Centralised Training -- in the case where the model trained on all the federation's private data is publicly exposed -- are also shown.

\noindent According to the literature the Model Inversion attack is very effective against the FedAvg. The global model contains information derived from the actual data of all participants, making it possible to extract individual data using inversion techniques~\cite{fredrikson2015model}. The same is true for the Membership Inference attack, which in a federated context can lead not only to identifying whether or not a piece of data is present in the model or not, but possibly also to guessing the federated member from which it originated~\cite{shokri2017membership}.
With FedKR, these attacks lose their effectiveness: even if an adversary managed to invert the model or infer the membership of a data sample, they would only gain information about the synthetic data, leaving the privacy of the real data intact. Resistance to Membership Inference attacks has also been demonstrated for models trained using a Knowledge Recycling technique, as in this case~\cite{lomurno2024synthetic}.

\noindent A similar reasoning can be extended to Gradients Leakage and Parameters Leakage attacks. Indeed, reconstructing data from parameters or updates would lead to synthetic data at best. However, since there is no global model and no exchange of gradients or parameters, such attacks are inherently ineffective.

\noindent The resilience benefits of FedKR over FedAvg remain even when more sophisticated attacks such as Data Poisoning or Byzantine and Sybil attacks are considered. FedKR's Dynamic Datasets Aggregation technique for data selection, combined with the synthetic nature of the data itself, provides a defence against such manipulation attempts. While in FedAvg a malicious participant can potentially significantly influence the overall model, in FedKR such attempts are easily identified and mitigated by filtering out suspicious or low quality contributions.

\noindent With respect to the Evasion attack, all models are partially vulnerable unless ad hoc countermeasures are applied. However, FedKR may offer a slight advantage if the synthetic data generated covers a wider space of possible inputs, potentially making the model more robust.

\noindent FedKR offers some advantages in dealing with Free-Riding attacks, but does not completely eliminate the problem. Although the quality of synthetic data can be assessed, a participant could theoretically still contribute with low quality or unrepresentative synthetic data. However, the nature of the generation and sharing process in FedKR makes it easier to identify and potentially exclude low quality contributions than in FedAvg.
With respect to Property Inference attacks, FedKR provides some protection by masking the exact statistical properties of the original data with synthetic data sampled from Gaussian distributions with user-defined standard deviations. However, it cannot be claimed that this provides complete protection, as a sophisticated attacker may still be able to infer some general properties of the original dataset by analysing the shared synthetic data.

\noindent FedKR also offers greater protection against Temporal attacks than FedAvg, as there is no need to exchange real-time model updates, which can be exploited by the attacker. However, it cannot be claimed that this attack surface is completely eliminated. Temporal differences in synthetic data generation and local training could still potentially reveal some partial information.

\section{Conclusions}\label{sec:Conclusion}
\noindent This paper introduced the Federated Knowledge Recycling (FedKR) technique, an innovative approach to federated learning applied in a cross-silo context. FedKR is based solely on the exchange of synthetic data between participants, eliminating the need to share potentially sensitive models, parameters or metadata.
Experimental results show that FedKR offers a favourable trade-off between performance and privacy, with an average accuracy improvement of 4.24\% compared to local training, proving particularly effective in data-poor scenarios such as the medical domain.
Due to the synthetic nature of the data exchanged, FedKR proved robust against attacks such as Model Inversion, Membership Inference and Gradients or Parameters Leakage, rendering them ineffective. In addition, the system provides significant mitigation against several other types of privacy attacks common to traditional federated learning systems.
The Dynamic Datasets Aggregation technique on which FedKR is based has proven effective in optimising the use of shared synthetic data, contributing to the overall performance of the system and providing an additional layer of protection. This methodology allows each participant to benefit from the collective knowledge of the federation while maintaining a high standard of privacy.

\section{Acknowledgements}\label{sec:Acknowledgements}
\noindent This paper is supported by the FAIR (Future Artificial Intelligence Research) project, funded by the NextGenerationEU program within the PNRR-PE-AI scheme (M4C2, investment 1.3, line on Artificial Intelligence).

\bibliographystyle{unsrt}  
\bibliography{references}  
\end{document}